%% file: PaperForReview.tex
\documentclass[10pt,twocolumn,letterpaper]{article}

\usepackage{cvpr}              

\usepackage{graphicx}
\usepackage{amsmath}
\usepackage{amssymb}
\usepackage{booktabs}

\newcommand{\norm}[1]{\left\lVert#1\right\rVert}%

\usepackage[pagebackref,breaklinks,colorlinks]{hyperref}

\usepackage[capitalize]{cleveref}
\crefname{section}{Sec.}{Secs.}
\Crefname{section}{Section}{Sections}
\Crefname{table}{Table}{Tables}
\crefname{table}{Tab.}{Tabs.}

\def\cvprPaperID{*****} 
\def\confName{CVPR}
\def\confYear{2022}

\begin{document}

\title{Towards a Unified View of Affinity-Based Knowledge Distillation}

\author{
Vladimir Li \quad Atsuto Maki\\
Division of Robotics, Perception, and Learning\\ KTH Royal Institute of Technology, Sweden\\
{\tt\small \{vlali, atsuto\}@kth.se}
}
\maketitle

\begin{abstract}
Knowledge transfer between artificial neural networks has become an important topic in deep learning. 
Among the open questions are what kind of knowledge needs to be preserved for the transfer, and how it can be effectively achieved. 
Several recent work have shown good performance of distillation methods using relation-based knowledge. 
These algorithms are extremely attractive in that they are based on simple inter-sample similarities. 
Nevertheless, a proper metric of affinity and use of it in this context is far from well understood. 
In this paper, by explicitly modularising knowledge distillation into a framework of three components, i.e. affinity, normalisation, and loss, we give a unified treatment of these algorithms as well as study a number of unexplored combinations of the modules.
With this framework we perform extensive evaluations of numerous distillation objectives for image classification,  
and obtain a few useful insights for effective design choices while demonstrating 
how relation-based knowledge distillation could achieve comparable performance to the state of the art in spite of the simplicity. 
\end{abstract}


\input{sections/introduction_rev}
\input{sections/related_work_rev}
\input{sections/method_rev}

\input{sections/results_rev}
\input{sections/discussion_rev}
\input{sections/conclusion}

\vspace{5mm}
\noindent
{\bf Acknowledgement} The support for this research by the Swedish Research Council through grant agreement no. 2016-04022 is gratefully acknowledged.

{\small
\bibliographystyle{ieee_fullname}
\bibliography{egbib}
}

\clearpage
\onecolumn
\appendix
\section{The ablation study in Section 4.4} 
This appendix provides complementary information regarding the ablation study described in Section 4.4. 

As part of the ablation study the best performing $\lambda$ for each teacher student (T-S) pair was found in a random search as shown on the right hand side of figure 5 (of the main paper). Results of the random search are shown below in figure \ref{fig:random_search} where each sub-figure corresponds to each T-S pair. For each T-S pair, we ran 40 experiments on CIFAR100 dataset using mAKD($\tt{CS\ L2\ SL1}$) as the distillation objective. We randomly sampled $\lambda$ from a log-uniform distribution in range $[e^7, e^{12}]$. A double exponential function (in dotted line) was fit to the final validation cross entropy loss values and the best $\lambda$ is picked at the minimum of that function.
\begin{figure}[h]
\centering
\begin{subfigure}{.48\textwidth}
  \centering
  \includegraphics[width=\linewidth,trim=6 7 7 -10, clip]{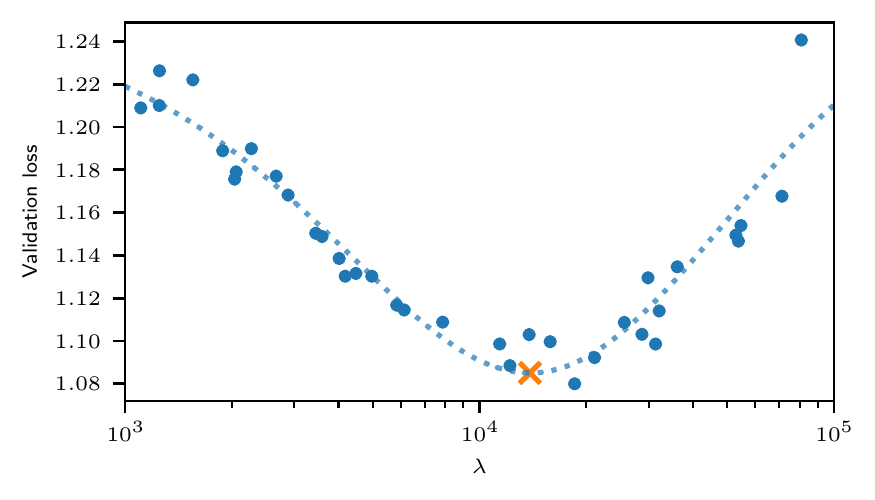}
  \caption{wrn-40-2 $\to$ wrn-40-1}
\end{subfigure}
\begin{subfigure}{.48\textwidth}
  \centering
  \includegraphics[width=\linewidth,trim=6 7 7 -10, clip]{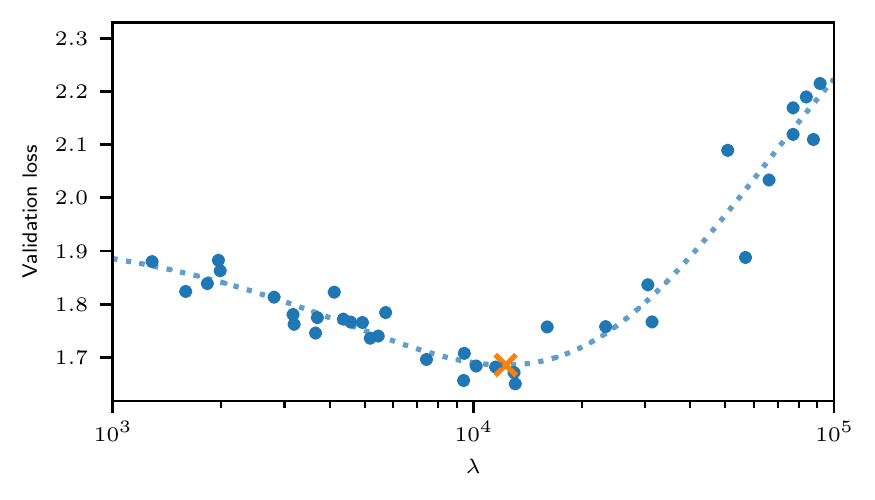}
  \caption{ResNet50 $\to$ MobileNetV2}
\end{subfigure}%

\begin{subfigure}{.48\textwidth}
  \centering
  \includegraphics[width=\linewidth,trim=6 7 7 -10, clip]{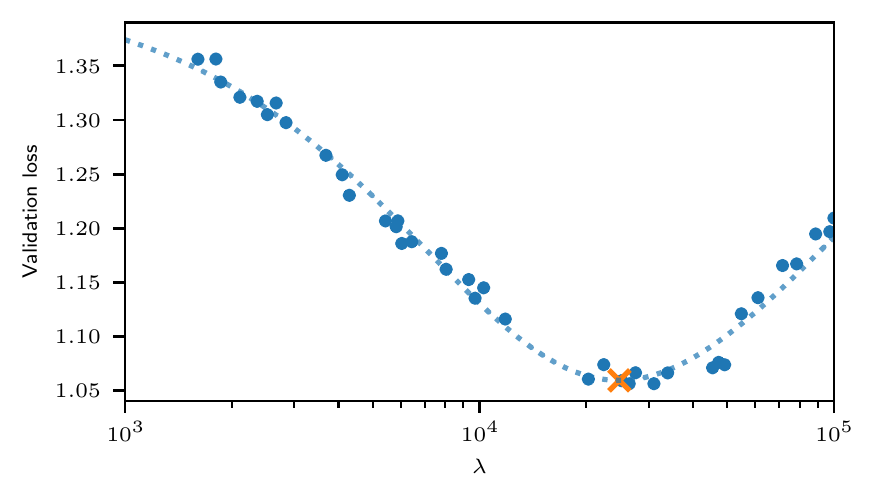}
  \caption{ResNet50 $\to$ vgg8}
\end{subfigure}
\begin{subfigure}{.48\textwidth}
  \centering
  \includegraphics[width=\linewidth,trim=6 7 7 -10, clip]{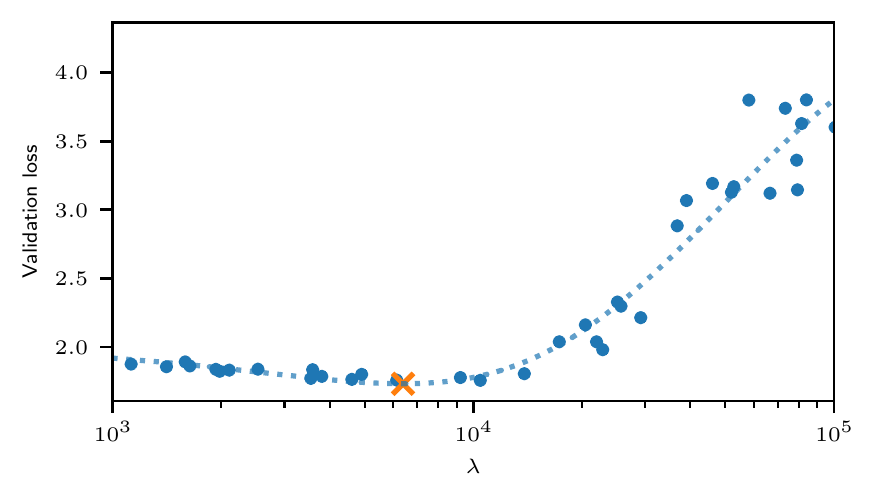}
  \caption{vgg13 $\to$ MobileNetV2}
\end{subfigure}

\begin{subfigure}{.48\textwidth}
  \centering
  \includegraphics[width=\linewidth,trim=6 7 7 -10, clip]{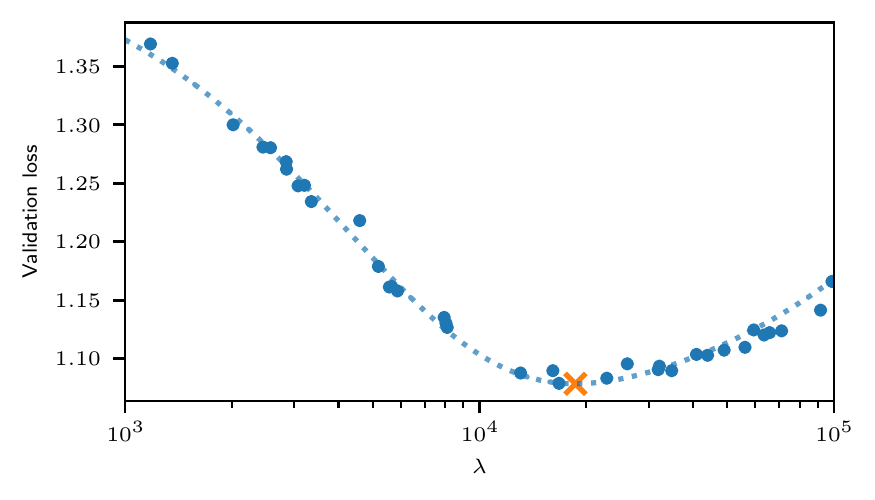}
  \caption{vgg13 $\to$ vgg8}
\end{subfigure}
\begin{subfigure}{.48\textwidth}
  \centering
  \includegraphics[width=\linewidth,trim=6 7 7 -10, clip]{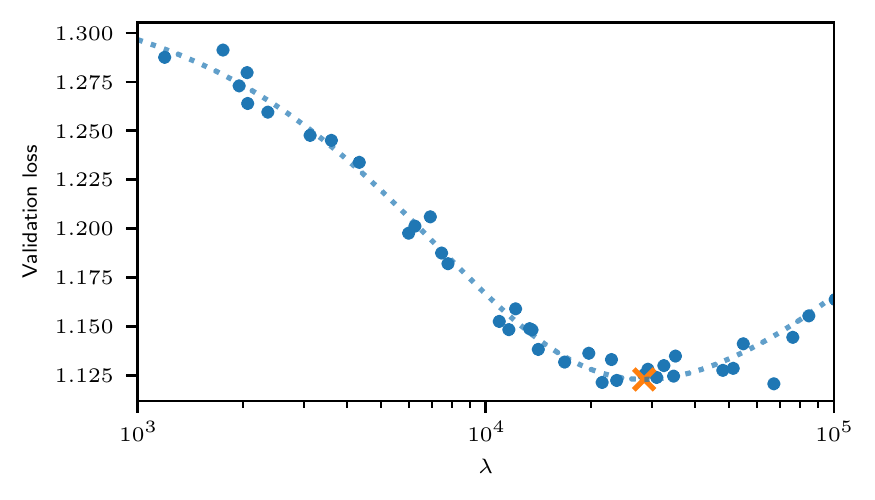}
  \caption{resnet56 $\to$ resnet20}
\end{subfigure}
\caption{Cross entropy loss for the validation set in a random search over $\lambda$. 
Each scatter point corresponds to one run. A double exponential function is fit to the data-points, in blue dotted line. The minimum of the function, orange cross, is set as the best $\lambda$ used in the ablation study. The caption under each plot corresponds to "teacher $\to$ student" pair.}
\label{fig:random_search}
\end{figure}

\begin{figure}  \ContinuedFloat
\begin{subfigure}{.48\textwidth}
  \centering
  \includegraphics[width=\linewidth,trim=6 7 7 -10, clip]{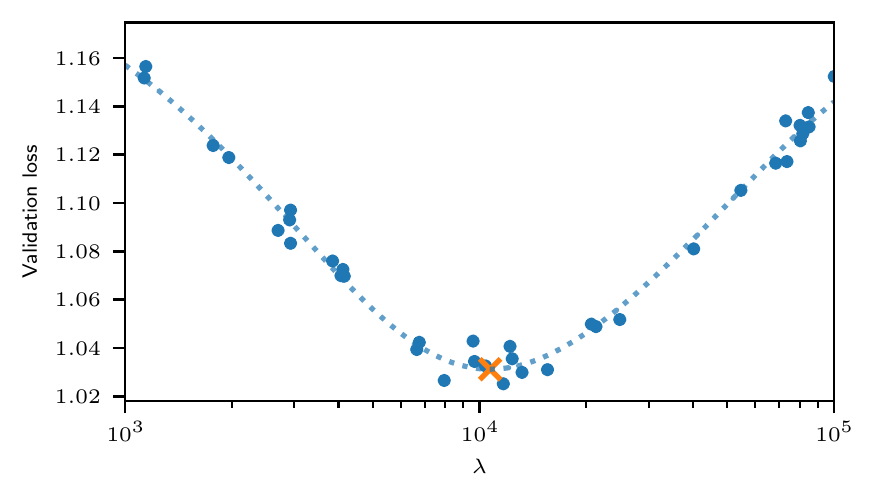}
  \caption{resnet32x4 $\to$ resnet8x4}
\end{subfigure}
\begin{subfigure}{.48\textwidth}
  \centering
  \includegraphics[width=\linewidth,trim=6 7 7 -10, clip]{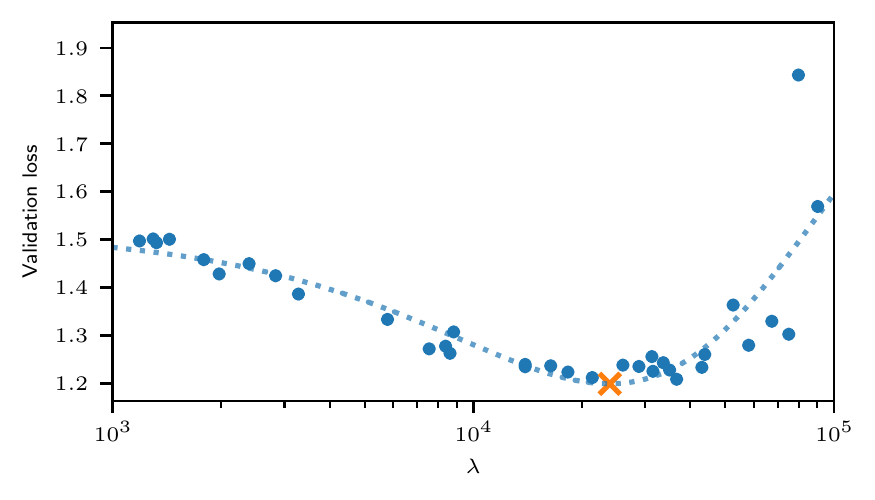}
  \caption{resnet32x4 $\to$ ShuffleNetV1}
\end{subfigure}

\begin{subfigure}{.48\textwidth}
  \centering
  \includegraphics[width=\linewidth,trim=6 7 7 -10, clip]{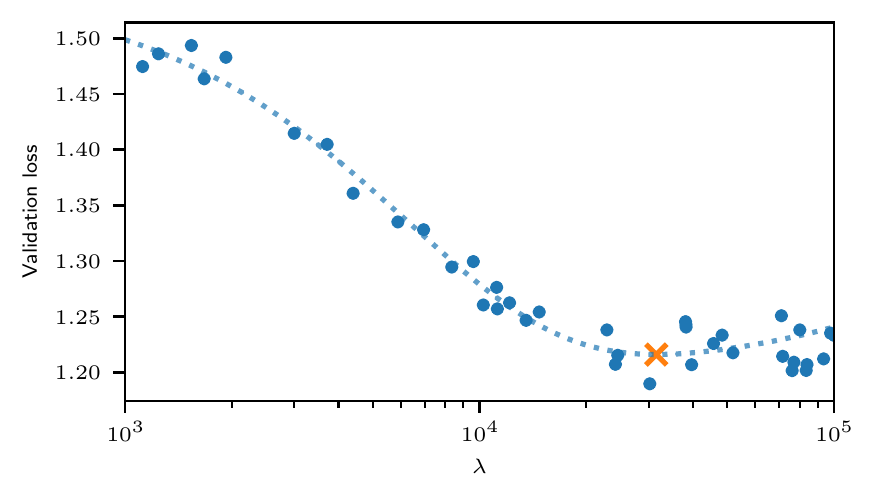}
  \caption{resnet32x4 $\to$ ShuffleNetV2}
\end{subfigure}
\begin{subfigure}{.48\textwidth}
  \centering
  \includegraphics[width=\linewidth,trim=6 7 7 -10, clip]{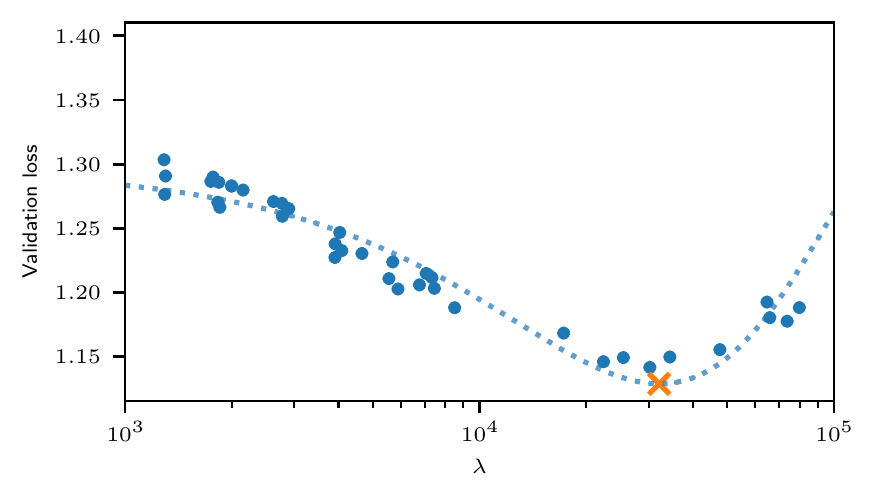}
  \caption{resnet110 $\to$ resnet20}
\end{subfigure}

\begin{subfigure}{.48\textwidth}
  \centering
  \includegraphics[width=\linewidth,trim=6 7 7 -10, clip]{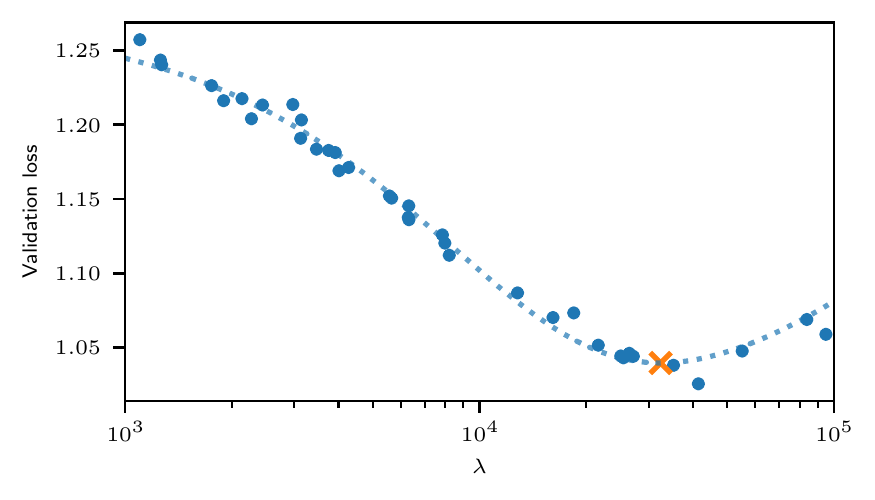}
  \caption{resnet110 $\to$ resnet32}
\end{subfigure}
\begin{subfigure}{.48\textwidth}
  \centering
  \includegraphics[width=\linewidth,trim=6 7 7 -10, clip]{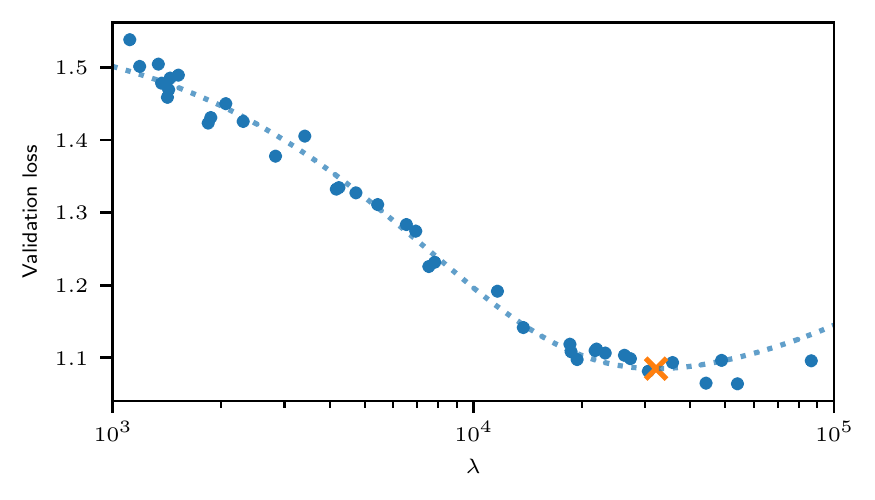}
  \caption{wrn-40-2 $\to$ ShuffleNetV1}
\end{subfigure}

\begin{subfigure}{.48\textwidth}
  \centering
  \includegraphics[width=\linewidth,trim=6 7 7 -10, clip]{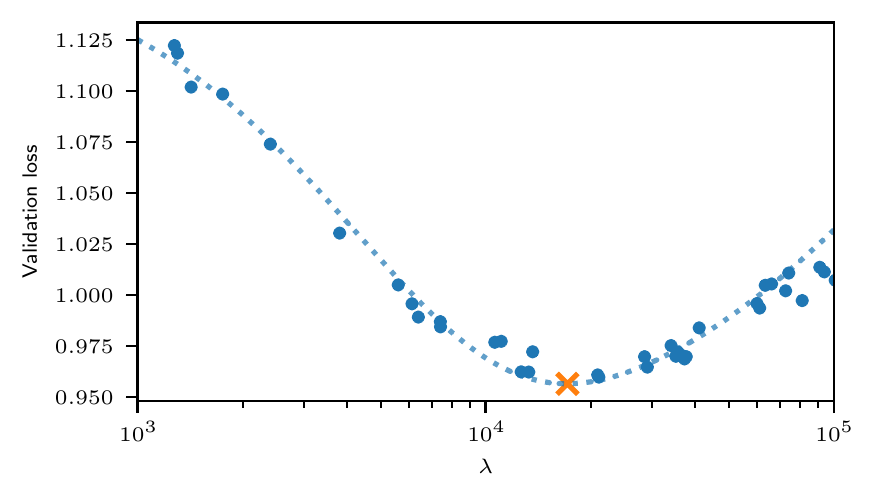}
  \caption{wrn-40-2 $\to$ wrn-16-2}
\end{subfigure}
\caption{See the caption on previous page.}
\end{figure}

\end{document}

%% file: sections/introduction_rev.tex
\section{Introduction}%
\label{sec:Introduction}%
Large-scale deep neural networks often have advantage in their performance over the smaller counterparts \cite{he2016deep, zagoruyko2016wide} thanks to their capacity. 
The size of networks can however be an issue in their deployment for production; due to a large memory footprint or substantial computational cost, those models may not necessarily fit well to small devices with limited power capacity, e.g.  a mobile phone.
While different approaches have been proposed to cope with this problem, such as designing effective architectures \cite{howard2017mobilenets, zhang2018shufflenet}, model pruning \cite{han2015learning, molchanov2016pruning}, low-rank approximation \cite{denton2014exploiting, jaderberg2014speeding}, and model quantization \cite{bulat2017binarized, lin2016fixed}, research based on knowledge transfer have shown promising directions in terms of knowledge distillation (KD), or distillation in short, which this paper is concerned with. 

The basic idea of knowledge transfer is to encode useful knowledge of  data by a teacher model and use that for training a smaller student network. Buciluǎ et al. \cite{bucilua2006model} were among the first that introduced knowledge transfer  in the context of artificial neural networks. They employed a large ensemble to label training data, and the labelled data were used to train a single neural network. Similar ideas were utilised in \cite{hinton2015distilling, ba2014deep, zagoruyko2016paying, ahn2019variational, romero2014fitnets} where a bulkier model was used to assist training of a smaller, shallower network. All these methods
use individual knowledge from single sample encoding/embedding which can be referred to as {\it instance-based} knowledge.

More recent work looked at distillation using inter-sample similarity, i.e. preserving the {\it relation-based} knowledge derived from intra-batch relationships among features of samples \cite{tung2019similarity, park2019relational, passalis2018learning, peng2019correlation}. Those methods are compelling as they exploit somewhat richer information, and have shown improved performance over instance-based approach. 
Yet, as observed in \cite{Ji2021S-A-Distill,tian2019contrastive,Yang2021KDSR}, they are largely outperformed by 
a few latest work including contrastive representation distillation. 
This could be due to the possible need for a more effective way for distilling the abundant information, including the challenge of defining proper distance between representations in the teacher and student networks \cite{chen2021W-CRD}.  
Indeed, given the considerable variation between the affinity-based methods, e.g. in terms of affinity measures, studies have been missing 
for motivating specific instances of relation-based KD \cite{Gou2021Survey}. 

\begin{figure}[t]
    \centering
    \includegraphics[width=0.33\textwidth,trim=25 13 9 15, clip]{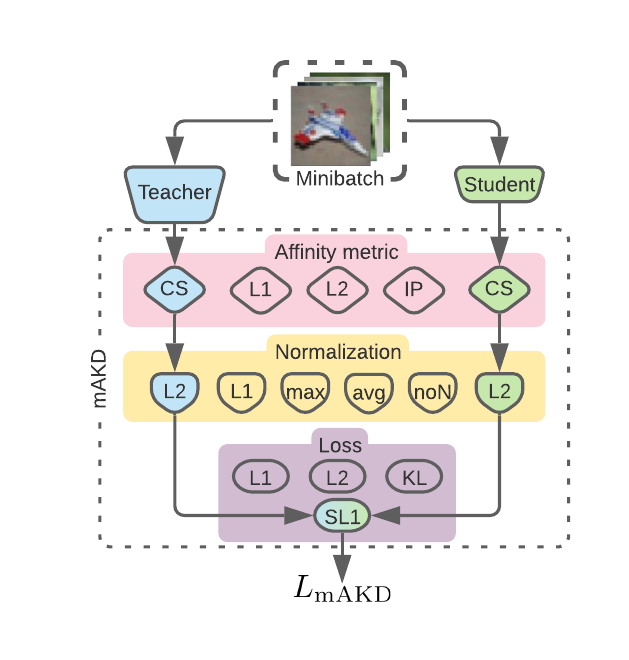}
    \caption{A schematic of the introduced framework, modular Affinity-based Knowledge Distillation (mAKD), defining a new objective $L_{\text{mAKD}}$. It consists of three components; affinity, normalisation, and loss. Each component has modules, described in section \ref{subsec:CombinatorialRelationalKnowledgeDistillation}, that can be swapped independently. 
    }
    \vspace{-6mm}
    \label{fig:banner}
\end{figure}

The primary goal of this paper is to explore this question; we study a wide variation of relation-based KD, discover effective variants, and demonstrate their performance compared to the state of the art. To this end, we perform a systematic study of putting possible variants in a modular setup, also including most of the existing relation-based knowledge distillation as its subset. Such a framework allows us to perform a rigorous comparison between different variants in a controlled setting.
We call our framework a modular Affinity-based Knowledge Distillation (mAKD) which is composed of three components. See figure \ref{fig:banner} for a sketch. Each component's modules can independently be swapped, resulting in numerous possible combinations (we consider 80 of them in our experiments). Using mAKD, we conduct a set of empirical experiments and identify well performing combinations as well as certain variants to avoid while providing important insights for them.

It should be noted that it is a challenging task to make fair assessment on numerous variants with different distillation objectives, i.e. different learning dynamics.
Our auxiliary goal is hence to find a meaningful balance between the main objective, i.e. cross-entropy loss for image classification in our case, and the knowledge distillation objective.
Although the interplay between the two is generally complex and not well understood, here we employ a strategy which we call GradNorm ratio preservation (GNoRP), inspired by the GradNorm \cite{chen2018gradnorm}. The aim of the GradNorm in the case of multitask learning setting is to balance the losses so that the training rate for each task is kept equivalent; the balancing is achieved by adjusting weighting factors, one for each task loss, such that the discrepancy between the gradient norms gets decreased. 
However,
GNoRP only requires a single weighting parameter, $\lambda$, which we dynamically adjust in such a way that the ratio between the gradient norms of the two losses remains constant. Such adaptation of GradNorm simplifies the method and allows for an interpretability of the GradNorm ratio.





Among a few state-of-the-art methods \cite{Ji2021S-A-Distill,tian2019contrastive,Yang2021KDSR} with respect to the performance, contrastive representation distillation (CRD) \cite{tian2019contrastive} was the first to outperform all the previous relation-based KD through a comprehensive comparison with multiple distillation losses on a large variety of teacher-student pairs -- those pairs commonly came into use in other benchmarks \cite{Ji2021S-A-Distill,Yang2021KDSR}.
Thus, 
we employ the same set of teacher-student pairs while ensuring to use common parameter settings, and choose to focus on the performance of CRD for our reference as well.
We then demonstrate that well-performing mAKD variants are able to achieve 
comparable performance. 
Note that this indicates the potential benefit of relation-based KD methods in the sense that they do not need a memory bank; CRD does require that for storing negative samples by the definition of its objective.

The main contributions are summarised as follows. 
(i) We present a modular framework for affinity-based knowledge distillation (mAKD) with which we give a unified treatment of related algorithms and explore numerous new variants, leading to guidelines for effective design choices.
(ii) A method for dynamically adjusting weighting of the additional loss: GradNorm ratio preservation (GNoRP).
(iii) The conclusions are based on solid experimental results, with insights for the effect, and will benefit the understanding of affinity-based knowledge distillation in general.


%% file: sections/related_work_rev.tex
\section{Related work}

Among the key factors in knowledge distillation is how to extract and then transfer the knowledge. Although there is still no consensus as to what sort of knowledge need be exactly transferred, two types of knowledge have been widely considered:
$a.$ instance-based knowledge from individual samples, i.e. the outputs of a specific layer (whether logits or latent features), and $b.$ relation-based knowledge which involves relationships among representations of samples; 
we concern ourselves with the second type in this paper, and refer to the relationships as affinity in the sequel. 
Our focus is on how to exploit affinities, while the introduced mAKD is inspired by a few recent work \cite{park2019relational, passalis2018learning, peng2019correlation, tung2019similarity} that explored this direction. 

In \cite{tung2019similarity}, the affinities between pairs of two sample representations were computed by their inner product, and stored in an affinity matrix for teacher and student network, respectively. Preceded by row-wise L2 normalisation, the mean squared error (MSE) was computed as the loss function between the teacher's and student's affinity matrices. 
In \cite{park2019relational}, two kinds of affinity metrics were introduced: distance-wise and angle-wise. The former is the Euclidean distance between every pair in the batch, normalised by the average of all the pairs, while the latter is to compute the cosine angle among every triplet in the batch. 
Huber loss was used as the function for measuring the discrepancy between the two affinity matrices.
In \cite{passalis2018learning}, shifted and scaled version of the cosine similarity metric was used. The shift and scale are such that the range is always constrained to $[0,1]$. They used row-wise L1 normalisation to ensure that each row becomes a valid conditional probability, and KL divergence as the loss function. 
In \cite{peng2019correlation}, three ways of measuring affinity were considered: naive MMD, bi-linear pool, and Gaussian RBF. The bi-linear pooling is equivalent to the inner product in \cite{tung2019similarity}. MSE was used as the loss function. 

Across those work based on relation-based knowledge, there is still a lack of consensus on the way to extract the affinity in the most effective fashion. 
In this respect, however, we identify a reoccurring composition of the proposed losses; they can be commonly regarded as consisting of an affinity-, normalisation-, and loss function. 
With those components
we define the framework of modular affinity-based knowledge distillation that allows extensive comparisons to be performed between various variants -- not only systematically but in a fair manner. 

%% file: sections/method_rev.tex
\label{sec:Method}%
\section{Method}%
This section consists of three parts, each accounting for
one of the contributions listed in section \ref{sec:Introduction}. First we describe components and modules that constitute the mAKD framework (section \ref{subsec:CombinatorialRelationalKnowledgeDistillation}), followed by the description of the combinatorial search (section \ref{subsec:Combinatorialsearch}) and GradNorm ratio preservation (section \ref{subsec:Dynamicbeta}). 

\subsection{Modular Affinity-based Knowledge Distillation}%
\label{subsec:CombinatorialRelationalKnowledgeDistillation}%
Given two ANN models, a pre-trained teacher $f^{(T)}$ and a randomly initialised student $f^{(S)}$, the goal of knowledge distillation is to train the student model with an assistance of the teacher model. A standard way to perform knowledge distillation is to define an additional loss function that is based on teacher's and student's representations. The main loss and a weighted KD loss are then jointly optimised using the loss function:%
\begin{equation}%
      \vspace{-1mm}
      L = L_{\text{main}} + \lambda \sum_{(l,l')} L_{\text{KD}}(z^{(S,l)}, z^{(T,l')})%
      \label{eq:weightedLoss}%
      \vspace{-1mm}
\end{equation}%
where $z^{(S,l)}$ and $z^{(T,l')}$ are feature representations of a batch produced by layer $l$ of the student and layer $l'$ of the teacher network, respectively. For simplicity, 
we consider the output of the penultimate layers as the feature representations. 
Without loss of generality, however, this can be extended to any layers in the teacher-student pair.%

The modular Affinity-based Knowledge Distillation (mAKD) objective consists of three main components: affinity metric $g$, normalisation of affinity matrix $f$, and a loss function $L$. $g$ is a measure of similarity/dissimilarity between samples within a batch. $f$ is used to relax the constraint of the mAKD objective in terms of the absolute magnitude of the affinity. 
$L$ defines the loss to minimise the discrepancy between the normalised affinity matrices from the student and teacher networks. 
Given the batch of penultimate feature representations from teacher $z^T$ and student $z^S$, the total loss is defined as:
\begin{equation}%
      L = L_{\text{main}} + \lambda L_{\text{mAKD}}(z^S, z^T)%
      \label{eq:totalLossFunc}%
\end{equation}%
where $L_{\text{mAKD}}$ is the mAKD objective:
\begin{equation}
    L_{\text{mAKD}}(z^S, z^T) = L(f(g(z^S)), f(g(z^T)))
    \label{eq:makd_objective}
\end{equation}
The concept in equation \ref{eq:makd_objective} is depicted in figure \ref{fig:banner}.

\subsubsection{Affinity metric}%
\label{ssubsec:Affinitymetric}%
Different pairwise affinity metrics have been proposed in the literature. For this study we consider affinity metrics according to two conditions: $a.$ simplicity of the metric and $b.$ the metric does not introduce additional parameters. A similarity score for every pair of samples in the batch is assigned in $g$. It constitutes an affinity matrix $G$ whose element $G_{i,j}$ corresponds to the affinity score between sample $i$ and sample $j$, $G_{i,j} = g(z_i, z_j)$. Note that $G$ is computed independently for teacher and student.%

- \textbf{Norm based affinity:} The simplest measures of affinity are the Manhattan and Euclidean distance, referred to as L1 and L2, respectively. Given a pair of feature vectors $z_i$ and $z_j$, the two metrics are:%
\begin{alignat}{2}%
     g_{\tt{L1}}(z_{i}, z_{j}) & = \norm{z_{i}-z_{j}}_1 \\%
     g_{\tt{L2}}(z_{i}, z_{j}) & = \norm{z_{i}-z_{j}}_2 %
\end{alignat}%

- \textbf{Inner product based affinity:} 
We consider the Inner Product between the two vectors ($\tt{IP}$) as well as Cosine Similarity ($\tt{CS}$) as candidates for the affinity metric.%
\begin{alignat}{2}%
     g_{\tt{IP}}(z_{i}, z_{j}) & = \sum_k z_{i,k}z_{j,k}                                          \\%
     g_{\tt{CS}}(z_{i}, z_{j}) & = \frac{\sum_k z_{i,k}z_{j,k}}{\norm{z_{i}}_2\norm{z_{j}}_2} %
\end{alignat}%

In total those make four affinity metrics: L1/L2 distances ($\tt{L1}/\tt{L2}$) and Inner Product ($\tt{IP}$), and Cosine Similarity ($\tt{CS}$).

\subsubsection{Normalisation}%
\label{ssubsec:Normalisation}%
A normalisation of $G$ can be performed in multiple ways. Let function $f$ represent a normalisation function, and the normalised affinity matrix $\tilde{G} = f(G)$.

- \textbf{Row-wise normalisation:} We can consider each row of $G$ as the affinity vector, $G_i$. $G_i$ conveys information about the similarity/dissimilarity of sample $i$ to all samples in the batch. Row-wise normalisation ensures that each affinity vector has a unit L1- or L2-norm:%
\begin{equation}%
      f_{\tt{L1}}(G) = \frac{G_{i,j}}{\norm{G_{i,:}}_1}, \hspace{3mm}  
      f_{\tt{L2}}(G) = \frac{G_{i,j}}{\norm{G_{i,:}}_2}.       
\end{equation}%

- \textbf{Matrix normalisation:} We can use the statistics of the entire affinity matrix, and consider average- and max-normalisation. In the former, the average across all elements in $G$ is made equal to one unit, whereas in the latter, the maximum value in $G$ is made equal to one unit:%
\begin{equation}
f_{\tt{avg}}(G)= \frac{b^2}{\sum_{i,j} G_{i,j}}G_{i,j}, \hspace{3mm} 
      f_{\tt{max}}(G) \frac{G_{i,j}}{\max(G_{i,j})}.
\end{equation}%
where $b$ is the batch size.

- \textbf{No normalisation ($\tt{noN}$):} If important information is hidden in the magnitude of the affinity metric we want to keep that information. Hence, not applying normalisation is a reasonable option.

In total we investigate five alternatives, i.e. L1/L2 row-wise normalisation ($\tt{L1}/\tt{L2}$), average/maximum ($\tt{avg}/\tt{max}$) matrix normalisation, and No normalisation ($\tt{noN}$).

\subsubsection{Loss function}%
\label{ssubsec:Lossfunction}%
The remaining component of mAKD is to compare the normalised affinity matrices from student and teacher networks. We define a loss such that the student learns to mimic inter-sample relations encoded by the affinity and normalisation functions. We investigate four different loss functions: three regression losses and KL divergence.

- \textbf{Regression losses:} The aim of the regression losses is to reduce the discrepancy between $\tilde{G}$'s.
We use L1/L2 distance, or a combination of both, i.e. SmoothL1 ($\tt{SL1}$) loss.%
\vspace{-2mm}
\begin{equation}
      L(\tilde{G}^{(S)}, \tilde{G}^{(T)}) = 
      \sum_{i,j} L_{\text{reg}}(\tilde{G}_{i,j}^{(S)}-\tilde{G}_{i,j}^{(T)})
\end{equation}
where $L_{\text{reg}}$ can be defined for given input $x$ as:
\vspace{-2mm}
\begin{alignat}{3}
     L_{\tt{L1}}(x)     & = \left|x\right|            \\
     L_{\tt{L2}}(x)     & = x^2                       \\
     L_{\tt{SL1}}(x)    & = \begin{cases}
            0.5x^2 & \text{if } x < 1.0 \\
            x-0.5  & \text{otherwise}
     \end{cases}
\end{alignat}

- \textbf{KL divergence:} 
To convert the affinity vectors into discrete probability distributions we use row-wise softmax before passing $\tilde{G}$ to KL divergence as the most common measure:%
\begin{alignat}{2}%
      L_{\tt{KL}}(\hat{G}^{(S)}, \hat{G}^{(T)}) & = %
      \dfrac{1}{b}
      \sum_{i,j} \hat{G}_{i,j}^{(T)}\log \dfrac{\hat{G}_{i,j}^{(T)}}{\hat{G}_{i,j}^{(S)}}
\end{alignat}%
where $\hat{G}_i = \text{softmax}(\tilde{G}_i)$.

In total we study four loss functions, i.e. L1/L2 distance ($\tt{L1}/\tt{L2}$), SmoothL1 loss ($\tt{SL1}$), and KL divergence ($\tt{KL}$). 

\subsubsection{mAKD loss and its connection to the related work}
The mAKD objective is constructed by putting the modules, as described in the previous sections, into the three components.  We expect the chosen variations (80 combinations) to well cover important combinations although other possible modules could be included. 
Here, every mAKD variant is conveniently identified by an acronym triplet, (affinity metric, normalisation method, loss function). For example, the objective function proposed in \cite{tung2019similarity} can be described as ($\tt{IP\ L2\ L2}$) while the distance-wise method in \cite{park2019relational} corresponds to ($\tt{L2\ avg\ SL1}$), and the bi-linear pooling version in \cite{peng2019correlation} to ($\tt{IP\ noN\ L2}$). In \cite{passalis2018learning}, shifted and scaled Cosine Similarity is used, paired with L1 normalisation and KL divergence. Due to the nature of shifting and scaling Cosine Similarity, there is no corresponding mAKD variant, however it is closely related to ($\tt{CS\ L1\ KL}$).

As a convention we will use the acronym in the sequel for specific variants of mAKD, e.g. mAKD($\tt{CS\ L1\ KL}$).
\subsection{Combinatorial search}%
\label{subsec:Combinatorialsearch}%
To find the most effective and meaningful combinations, we designed a combinatorial search. For a proper comparison between different mAKD variants we imposed the following requirements:

\begin{enumerate}
    \item The mAKD loss should be tested in isolation.
    
    \item The effect of any mAKD variant, on the update of the model, should be equivalent to each other. 
\end{enumerate}%
To conform the first requirement, we decomposed the model into two parts, a feature extractor (the entire model except the last classification layer) and a classifier. A variant of mAKD loss was used to train a feature extractor. To validate the effectiveness of the loss, we trained a linear classifier with the features extracted from the training set. We use the validation accuracy of the classifier as the performance metric for all the mAKD variants.%

The second requirement is fulfilled by weighting mAKD objective by a fixed $\lambda$. 
$\lambda$ is chosen such that the initial gradient norm of the mAKD is made equal to the average value of the initial gradient norm ($0.07$ units) that is observed when different randomly initialised models are trained with classification loss only. Note that the gradient norm is w.r.t. the penultimate feature vector $z$.
By applying such $\lambda$, we are able to ensure that different mAKD variants have equal importance on the update of the model parameters, resulting in a fairer comparison. %

Furthermore, we use the same set of hyper-parameters as used in the training with classification loss only. Doing so, it is safe to regard that 
additional 
distillation loss will have stable convergence when paired with the original loss.

\subsection{Gradient Norm Ratio Preservation (GNoRP)}%
\label{subsec:Dynamicbeta}%
It is natural to anticipate that the optimal weighting parameter $\lambda$ (see eq. \ref{eq:totalLossFunc}) can be different for different mAKD variants as well as for different teacher-student architecture pairs. In fact, during our experiments we observed varying impacts on the update of the parameters for different mAKD objectives. This was also the case with an identical mAKD variant applied to different teacher-student pairs. 

To deal with this issue, we propose a method for dynamic adjustment of $\lambda$ which is inspired by the GradNorm \cite{chen2018gradnorm}. Unlike GradNorm, however, we relax the objective and simply adjust $\lambda$, with every new mini-batch, so that the gradient norm ratio, $r_{GN}$, between the main classification criterion and the mAKD objective is kept constant. We compute the gradient norm with respect to the penultimate feature representation $z$. $\lambda$ is obtained by an  optimisation of the following loss function:
\begin{equation}
    L_{\text{GNoRP}} = 
    \left(
    r_{GN} \norm{\dfrac{\partial L_{\text{main}}}{\partial z}}_2 - \norm{\dfrac{\partial \lambda L_{\text{mAKD}}}{\partial z}}_2
    \right)^2
\end{equation}
where $L_{\text{main}}$ and $L_{\text{mAKD}}$ denote classification loss and mAKD loss, respectively. 


Using this formulation, we are able to examine different  values of $r_{GN}$ in order to find the one that works well across multiple settings.

%% file: sections/results_rev.tex
\section{Experiments}
We perform four groups of experiments. The first is to find a well-performing set of mAKD variants by a combinatorial search (section \ref{subsec:combinatorialsearchresults}). 
We then proceed with finding a proper gradient norm ratio $r_{GN}$ (section \ref{sec:alphasearch}). 
The results with mAKD loss are compared to those by one of the state-of-the-art methods, i.e. CRD loss, on CIFAR100 and ImageNet datasets (section \ref{sec:comarison_with_sota}). Finally, GNoRP is validated by comparing its performance against the case with the best performing $\lambda$ for each T-S pair (section \ref{sec:dynamicBetaAblation}).
Those are followed by discussions in section \ref{sec:discussion}.

\subsection{Combinatorial search}
\label{subsec:combinatorialsearchresults}
\begin{figure*}[h!]
    \centering
    \includegraphics[width=\linewidth, trim=5 1 5 2, clip]{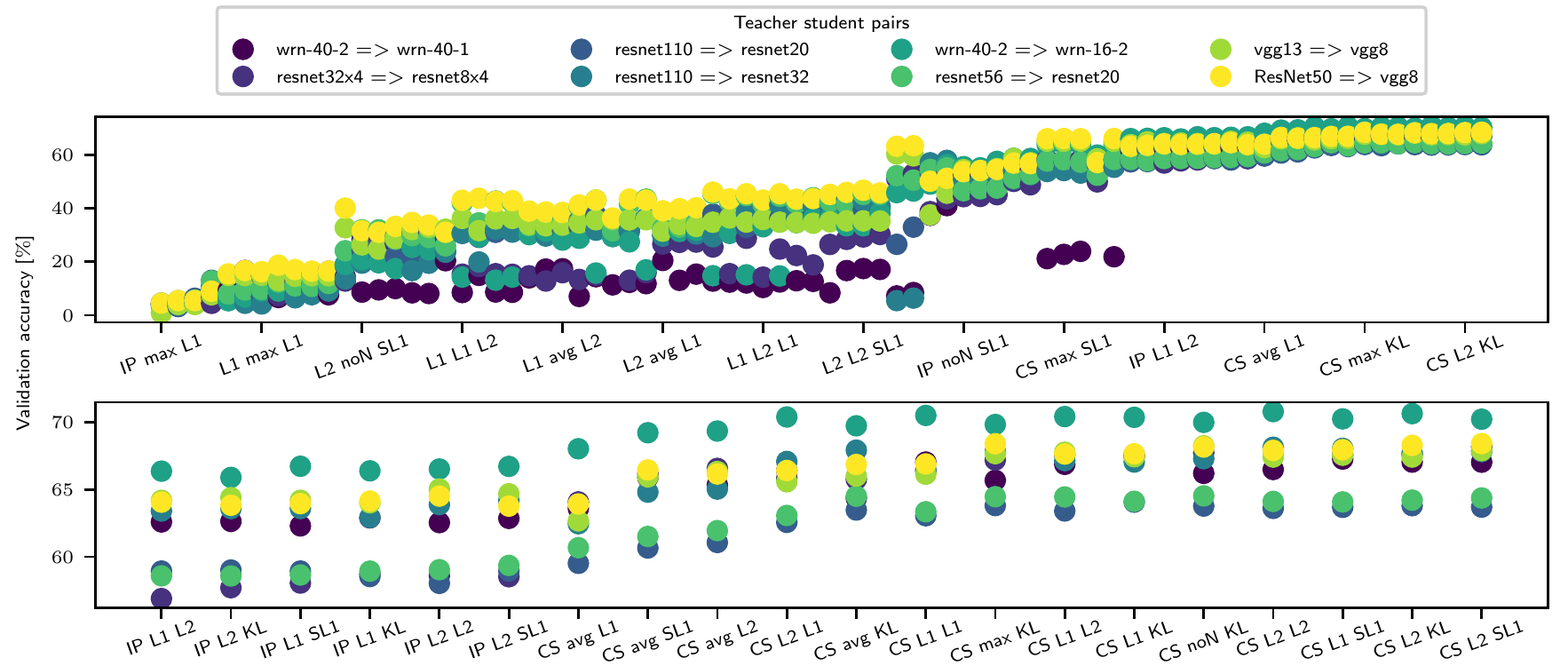}
    \vspace{-3mm}
    \caption{
    Results from the combinatorial search on CIFAR100. 
    Top figure for all mAKD variants and 20 best variants in the bottom.
    In both figures the horizontal-axis is for mAKD variants used to train a student model (not all the variants are indicated due to space in the top figure). The performance in terms of the validation accuracy (on the vertical-axis) was evaluated by training a linear classifier based on the features extracted from the model.
    See section \ref{subsec:CombinatorialRelationalKnowledgeDistillation} for the label abbreviations.
    }
    \vspace{7mm}
    \label{fig:initial_exp_full}
    \centering
    \includegraphics[width=\linewidth, trim=6 7 11 6, clip]{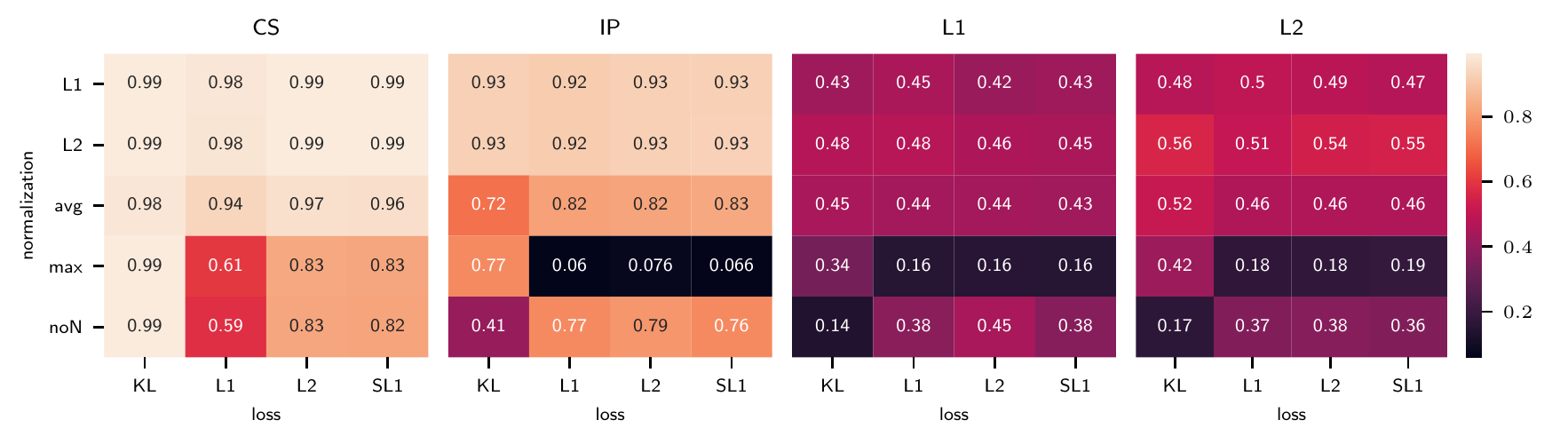}
    \vspace{-3mm}
    \caption{
    Summary of combinatorial search for the same experiments that are shown in figure \ref{fig:initial_exp_full}. 
    Each subplot corresponds to a specific affinity metric, each row to a normalisation function, and each column to a different loss function. 
    The value on each cell is the average of normalised validation accuracy.
    }
    \label{fig:initial_exp_color}
\end{figure*}

To find out the most meaningful combination of the mAKD loss we run eight experiments\footnote{Eight T-S pairs, which are the subset of the pairs used in \cite{tian2019contrastive}, as shown in the legend of figure \ref{fig:initial_exp_full}.}, for each of the 80 mAKD variants.
In figure \ref{fig:initial_exp_full} we report the validation accuracy (vertical-axis) for each T-S pair (different colours) and mAKD variants (along horizontal-axis). On the top plot we report all 80 mAKD variants sorted by the average accuracy, while top 20 best variants were reported on the plot in the bottom. From figure \ref{fig:initial_exp_full} we see that the $\tt{CS}$ affinity is consistently better than the rest. It is followed by $\tt{IP}$ affinity and other distance-based metrics. 

In a more intuitive visualisation over mAKD variants, in figure \ref{fig:initial_exp_color}, each cell reflects the average of normalised validation accuracy across different T-S pairs. The normalisation is performed such that the maximum accuracy for a T-S pair becomes equal to one. A few of our main observations based on figure \ref{fig:initial_exp_color} are: 

- $\tt{CS}$ affinity performs well across a wide range of normalisation-loss pairs, closely followed by $\tt{IP}$ affinity.

- $\tt{max}$ normalisation has significantly worse performance than others except the cases with $\tt{CS}$ affinity.

- $\tt{KL}$ loss works well for the $\tt{CS}$ affinity, across any normalisation functions.

- No normalisation with $\tt{KL}$ loss is consistently worse than any others for all affinity metrics except for $\tt{CS}$ affinity.

Based on those observations, in further experiments we choose to use Cosine Similarity, L2 normalisation and SmoothL1 loss, i.e. mAKD($\tt{CS\ L2\ SL1}$). However, we should point out that the performance across the best eleven mAKD variants in figure \ref{fig:initial_exp_full} is similar to each other and any of those variants would work as well as the single choice we made above.

\subsection{Optimal GradNorm ratio, $r_{GN}$}
\label{sec:alphasearch}
\begin{figure}[t]
    \centering
    \includegraphics[width=\linewidth, trim=6 6 6 6, clip]{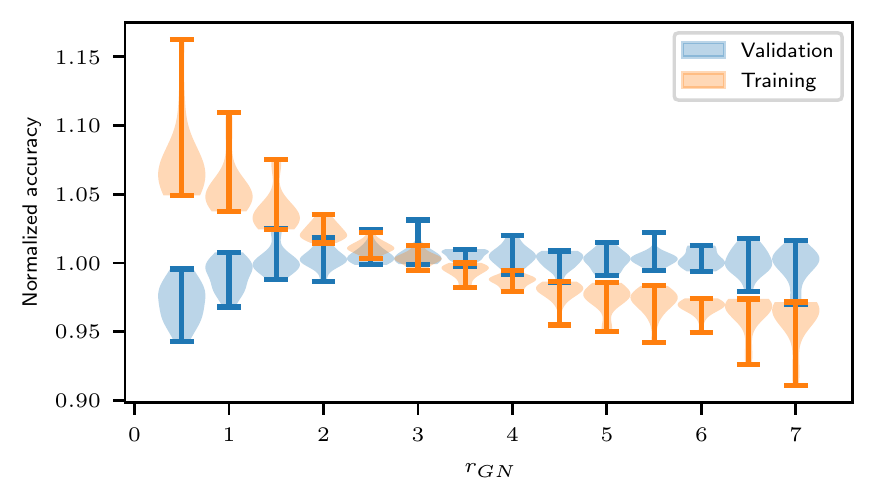}
    \vspace{-6mm}
    \caption{
    Normalised accuracy (along vertical-axis) for a range of $r_{GN}$ (horizontal-axis) as well as for different model pairs. 
    The performance (in blue) improved according to $r_{GN}$ up to $r_{GN}=3.5$. For $r_{GN}>3.5$ the validation accuracy plateaus while the variance is increased. 
    The training accuracy (orange) is decreased with increased $r_{GN}$.
    }
    \label{fig:alpha_search}
\end{figure}

In this experiment we investigate the effect of different $r_{GN}$, on CIFAR100. For every T-S pair, we run experiments by varying $r_{GN}$ from $0.5$ to $7$ with increment of $0.5$. The results are shown in figure \ref{fig:alpha_search}. Due to the variation in the accuracy across different T-S pairs, we perform a normalisation; the accuracy for each T-S pair is divided by the average across all $r_{GN}$ values. We can observe that the validation accuracy is increased with increasing $r_{GN}$, signifying the importance of the mAKD loss. The training accuracy is decreased with increasing $r_{GN}$, indicating that using mAKD reducing overfitting. Based on our observation we choose to use $r_{GN}=3.5$ in CIFAR100 experiments.

\subsection{Comparisons}
\label{sec:comarison_with_sota}

As explained earlier in section \ref{sec:Introduction}, for our benchmark we choose \cite{tian2019contrastive} among the state-of-the-art methods while ensuring to use common architectures and parameter settings. 

\textbf{CIFAR100:} For fair and strict comparisons we rerun the experiments\footnote{The test accuracy found in our tables are naturally slightly different from those reported in \cite{tian2019contrastive} after re-running the experiments rather than copying their results.}  performed with the CRD loss. In doing so we make sure that the pretrained teacher and random seeds are the same for CRD and mAKD experiments. For every T-S pair and each objective we had five runs with different random seeds. Note that the set of random seeds is kept the same. The teacher model is also kept the same across different runs. Table \ref{tab:crd_comparison} shows top-1 and top-5 test accuracy for CRD and mAKD ($\tt{CS\ L2\ SL1}$) losses. The performance of mAKD is comparable with CRD; while CRD is superior in terms of top-1 accuracy, mAKD has better performance in terms of top-5 accuracy. mAKD loss has especially good performance when the student model is vgg8 or one of the ShuffleNet versions. 

\begin{table*}
    \centering
    \input{figures/tables/comparison-v2.tex}
    \caption{Average test accuracy in using different T-S pairs (for CIFAR100) and its standard deviation across five independent runs with different seeds. For this comparison we use $r_{GN}=3.5$ and mAKD($\tt{CS\ L2\ SL1}$). Best top-1/top-5 accuracy between CRD and mAKD is marked with bold font.}
    \label{tab:crd_comparison}
\end{table*}
\begin{table}
    \centering
    \vspace{-2mm}
    \input{figures/tables/imNet_exp.tex}
    \caption{Test error for training using mAKD on the ImageNet dataset. $r_{GN} = 1$ and mAKD($\tt{CS\ L2\ SL1}$) were used. The CRD results are from \cite{tian2019contrastive}.}
    \label{tab:imnet_exp}
    \vspace{-5mm}
\end{table}

\noindent
\textbf{ImageNet:} We used the same setup as in \cite{tian2019contrastive}. Since it is costly to perform a search for optimal $r_{GN}$ for ImageNet dataset, an uninformative prior\footnote{Discussed in section \ref{sec:discussion}.} $r_{GN}=1$ is used. The results are in table \ref{tab:imnet_exp}. While the CRD loss has better top-1 error, the mAKD loss achieves better top-5 error.
We discuss this further in section 
\ref{sec:discussion}.

\subsection{Ablation study for the importance of GNoRP}
\label{sec:dynamicBetaAblation}
\begin{figure}[t]
        \centering
    \includegraphics[width=\linewidth]{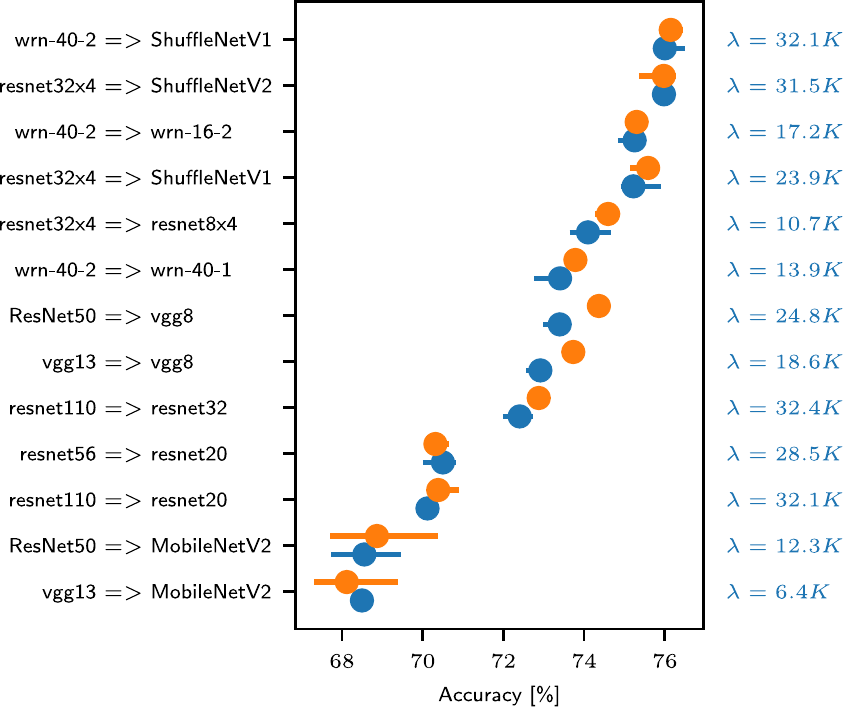}
    \vspace{-2mm}
    \caption{
    Performance with the static $\lambda$ (blue) and GradNorm ratio preservation (orange) on CIFAR100, using mAKD($\tt{CS\ L2\ SL1}$). 
    The average test accuracy plotted with min and max accuracy on the error bar. 
    The static $\lambda$ values are to the right of the plot.
    }
    \label{fig:beta_ablation}
\end{figure}

To verify the effect of the GradNorm ratio preservation, in this study we examined different values of static $\lambda$ during the training. For each T-S pair the best performing $\lambda$ is found in a random search (40 runs per T-S pair, resulting in 520 runs in total)\footnote{Results of the random search are in Appendix.}. Then the model is trained using the entire training set for five different random seeds
(the same set of seeds as for CIFAR100 experiments).
The results are in figure \ref{fig:beta_ablation}. We can see that the selected static $\lambda$, reported to the right of the plot, have a large variance across different T-S pairs, while the performance for the GNoRP with $r_{GN}=3.5$ (orange) is comparable, and often even better than the results with static $\lambda$ (blue). 

\subsection{Implementation details}
For all experiments we use hyper-parameters and teacher-student pairs as studied in \cite{tian2019contrastive}. For the experiments with CIFAR100 dataset \cite{krizhevsky2009learning} all teacher-student pairs can be seen in table \ref{tab:crd_comparison}. Batch size is set to 64 and weight decay to 0.0005. SGD with momentum of 0.9 and learning rate of 0.05 is used for all cases except for ShuffleNet \cite{zhang2018shufflenet} and MobileNet \cite{howard2017mobilenets} architectures, where learning rate of 0.01 is applied. The experiments were run for 240 epochs and the learning rate was dropped by factor of 10 at epoch 150, 180 and 210. For the faster search experiments (sections \ref{subsec:combinatorialsearchresults}, \ref{sec:alphasearch}, and random search in section \ref{sec:dynamicBetaAblation}), we chose to run those with decreased number of epochs. Those experiments are run for 80 epochs while learning rate is dropped at epoch 50, 60 and 70.

For the ImageNet \cite{deng2009imagenet} experiments we used a pretrained ResNet-34 \cite{he2016deep} as teacher and ResNet-18 as student model. Batch size is set to 256, initial learning rate to 0.1, weight decay to 1e-4 and momentum to 0.9. We train for 100 epochs and learning rate is dropped by factor of 10 at epoch 30, 60 and 90.

In all search experiments validation data was used for evaluation which was created by randomly sampling 40\% of the training data.

To avoid unexpected influence from GNoRP we employ
a separate optimiser for GNoRP (we use Adam with its default parameters in pyTorch) 
and
$\breve{\lambda} = \log(\lambda)$ as the variable for optimisation to keep $\lambda$ positive, i.e. we use $\lambda=\exp(\breve{\lambda})$.


For the evaluation of the mAKD loss in the combinatorial search (section \ref{subsec:combinatorialsearchresults}), we trained a linear classifier using a LinearSVC from scikit-learn \cite{scikit-learn}. All parameters of LinearSVC, except for dual, were kept as default. The dual was set to False in accordance with the recommendations in the documentation. Note that the GNoRP was not used in combinatorial search experiments.

%% file: figures/tables/comparison-v2.tex
\begin{tabular}{llllllll}
    \toprule
               &              & \multicolumn{3}{l}{top1} & \multicolumn{3}{l}{top5}                                                                                     \\
            Teacher(T)   & Student(S)             & T/S          & CRD                      & mAKD           & T/S & CRD                 & mAKD           \\
    \midrule
    ResNet50   & MobileNetV2  & 78.68/64.34              & \textbf{69.21$\pm$0.15}      & 68.87$\pm$0.87          & 95.32/87.67     & 90.87$\pm$0.26          & \textbf{91.49$\pm$0.34} \\
               & vgg8         & 78.68/70.28              & 74.02$\pm$0.22               & \textbf{74.37$\pm$0.09} & 95.32/90.29     & 92.86$\pm$0.05          & \textbf{93.37$\pm$0.16} \\
    resnet110  & resnet20     & 73.90/69.30              & \textbf{71.24$\pm$0.10}      & 70.39$\pm$0.29          & 92.92/91.89     & \textbf{92.51$\pm$0.12} & 92.42$\pm$0.06          \\
               & resnet32     & 73.90/71.00              & \textbf{73.27$\pm$0.30}      & 72.88$\pm$0.12          & 92.92/92.53     & 93.03$\pm$0.16          & \textbf{93.25$\pm$0.07} \\
    resnet32x4 & ShuffleNetV1 & 79.06/71.55              & 75.28$\pm$0.19               & \textbf{75.59$\pm$0.24} & 94.64/91.20     & 93.29$\pm$0.13          & \textbf{93.86$\pm$0.12} \\
               & ShuffleNetV2 & 79.06/72.42              & 75.77$\pm$0.36               & \textbf{75.97$\pm$0.32} & 94.64/91.67     & 93.73$\pm$0.23          & \textbf{94.02$\pm$0.14} \\
               & resnet8x4    & 79.06/72.54              & \textbf{75.37$\pm$0.04}      & 74.60$\pm$0.19          & 94.64/92.44     & \textbf{94.02$\pm$0.16} & 93.96$\pm$0.19          \\
    resnet56   & resnet20     & 72.72/69.30              & \textbf{71.18$\pm$0.25}      & 70.32$\pm$0.25          & 92.35/91.89     & 92.46$\pm$0.23          & \textbf{92.73$\pm$0.14} \\
    vgg13      & MobileNetV2  & 74.88/64.34              & \textbf{69.07$\pm$0.16}      & 68.12$\pm$0.73          & 92.73/87.67     & \textbf{91.22$\pm$0.14} & 90.99$\pm$0.32          \\
               & vgg8         & 74.88/70.28              & 73.42$\pm$0.30               & \textbf{73.73$\pm$0.15} & 92.73/90.29     & 92.67$\pm$0.17          & \textbf{92.72$\pm$0.18} \\
    wrn-40-2   & ShuffleNetV1 & 76.17/71.55              & 75.68$\pm$0.20               & \textbf{76.15$\pm$0.22} & 93.55/91.20     & 93.40$\pm$0.06          & \textbf{94.07$\pm$0.11} \\
               & wrn-16-2     & 76.17/73.44              & \textbf{75.53$\pm$0.35}      & 75.30$\pm$0.18          & 93.55/93.05     & 93.99$\pm$0.11          & \textbf{94.21$\pm$0.07} \\
               & wrn-40-1     & 76.17/71.39              & \textbf{74.09$\pm$0.27}      & 73.78$\pm$0.15          & 93.55/92.40     & 92.93$\pm$0.20          & \textbf{93.54$\pm$0.07} \\
    \bottomrule
\end{tabular}

%% file: figures/tables/imNet_exp.tex
\begin{tabular}{lrrrrrrrr}
\toprule
{} &  Teacher &  Student &  CRD &   mAKD \\
\midrule
Top-1 &    26.69 &    30.25 &  \textbf{28.83} &  29.14 \\
Top-5 &     8.58 &    10.93 &  9.87 &  \textbf{9.78} \\
\bottomrule
\end{tabular}

%% file: sections/discussion_rev.tex
\section{Discussion}
\label{sec:discussion}

\hspace*{4mm}
\textbf{For affinity inner product based metrics with a merit.} 
It is observed that the choice of the affinity module is a dominating factor in constructing a well-performing mAKD objective from the results of our experiments for image classification. We saw a considerable merit in using inner product based affinity functions, i.e. $\tt{CS}$ and $\tt{IP}$, as opposed to $\tt{L1}/\tt{L2}$,  in the setup of using the penultimate representation. 
The advantage can be explained by the good alignment between the inner product based affinity module and the operations performed in the classification layer. 
That is, while the classification layer computes the inner product of the sample feature and class representation, those affinity modules also compute the inner product of the sample representations within a batch.
\vspace{1mm}


\textbf{Normalisation not critical with cosine similarity.}
For the normalisation part, some available observations are: 
$a.$ normalisation generally helps, $b.$ max normalisation should be avoided, and $c.$ the choice is less important for cosine similarity affinity module. 

$a.$ 
The compactness of the student model implies its lower capacity, 
making it usually hard to mimic the manifold of teacher's representation. By applying normalisation, however, 
such a restriction is expected to be softened.

$b.$ 
$\tt{max}$ matrix normalisation performs generally poorly. This may well be due to the imbalance among the affinity scores for different pairs of samples, i.e. an extremely high affinity for a certain pair would cause loss of information, making most of the other entries negligibly small.

$c.$ $\tt{CS}$ achieves good performance across a diverse set of normalisation functions even including $\tt{max}$ matrix normalisation. This is explained by virtue of internal normalisation of the cosine similarity metric.
\vspace{1mm}

\textbf{The choice of loss function has less impact.}
When we pick row-wise normalisation or normalisation by average, the choice of loss function has smaller impact on the performance of the mAKD objective. To be noted is that KL divergence benefits from any normalisation. This is likely thanks to a similar effect to that by the temperature scaling in the original knowledge distillation \cite{hinton2015distilling}, where normalisation factor (inverse temperature) smaller than one has a smoothing effect on softmax distribution, which helps recover the hidden affinity between sample pairs. 
\vspace{1mm}


\textbf{Benefits of GNoRP.} There are two benefits in using GNoRP instead of static weighting: $a.$ it {\it absorbs} the differences that occur due to different architectures or different definitions of the loss functions and $b.$ it allows for better interpretability of the GradNorm ratio $r_{GN}$, 
than
the weighting parameter $\lambda$. Both benefits are related to the fact that controlling GradNorm ratio has a more direct effect on the update of model's parameters than a static weighting of the losses. Since the gradient norm is approximately proportional to the step made in the parameter space, the GradNorm ratio $r_{GN}$ would correspond to the ratio of step sizes from the two losses. For example, setting $r_{GN} = 2$ implies that the step size due to the mAKD loss is twice as large as that due to the main loss. Following this explanation we can interpret the gradient norm ratio as the importance ratio between the losses. Furthermore, if a hyper-parameter search is too expensive, one could use an uninformative prior $r_{GN}=1$, signifying that the two losses are given an equal importance. 
\vspace{1mm}

\textbf{The top-5 accuracy.}
The mAKD loss tends to give better performance in terms of top-5 accuracy, in particular, with mAKD($\tt{CS}$ $\tt{L2}$ $\tt{KL}$) as seen in section \ref{sec:comarison_with_sota}. 
Our interpretation is that affinity-based KD can be better suited to capture the class similarity intrinsic to given training data, e.g. {\it cat} is a closer category to that of {\it dog} than {\it bus}.
Although this is rather intuitive, it could bring further benefit in open set classification problem settings where a sample that does not have a clear association to one single class, but it could be explained as a combination of multiple categories.
\vspace{1mm}


\textbf{Applications of mAKD framework.}
We have investigated a variety of affinity based objectives, thus far on image classification tasks. If one would like to utilise mAKD framework in a similar setup the recommendation is to use one of the best performing variants as identified in figure \ref{fig:initial_exp_color}. When using affinity-based knowledge distillation in a different scenario one could construct another combinatorial search. Without loss of generality one can reduce the search space to the following modules: $\tt{CS}$ and $\tt{L2}$ for affinity metric, $\tt{L2}$ normalisation, and $\tt{KL}$ and $\tt{SL1}$ as loss functions. 
However, the framework is open for potential extension with other modules for the task at hand.

%% file: sections/conclusion.tex
\section{Conclusions}
For better understanding affinity-based knowledge distillation methods, in this paper we presented a modular framework called mAKD, composed of three components. 
Through a set of systematic evaluations using mAKD, 
we gave a unified treatment of related algorithms while exploring 80 variants using different modules, and led to guidelines for effective design choices for image classification.
We also introduced GradNorm ratio preservation (GNoRP) method for dynamic weighting of the distillation loss which allows for interpretability of the balance against the main objective. 
The obtained insights are all based on solid experiments.
The experimental results show that well-designed variants could attain, in spite of their simplicity, a similar level of performance to those of state-of-the-art methods, although the main takeaway of this paper is rather in the insight gained on relation-based knowledge transfer through the unified view.

We hope that research into further challenges for designing effective distillation, e.g. in 
\cite{Zhu2021C-RCD,Ji2021M-Pairwise,Yuan2020_RevisitKD,Li2020LocalKD,Xu2020SSKD,Lan2018-KD_otf}, 
will also benefit from the observations gained through different variants in the presented framework.




